\documentclass[10pt]{article}
\usepackage[a4paper,margin=1.4in]{geometry}
\usepackage{bbm} %
\usepackage{pdfsync,amssymb,amsmath,ifpdf,color,placeins}
\usepackage{amssymb}
\usepackage{amsmath}
\usepackage{multirow}
\usepackage{graphicx}
\usepackage{booktabs}
\usepackage{caption} 
\usepackage[dvipsnames]{xcolor}

\definecolor{refcolor}{rgb}{0,0,0.5}
\ifpdf
\PassOptionsToPackage{hyphens}{url}
\usepackage[%
  pdftitle={Surpassing state of the art on AMD area estimation from RGB fundus images through careful selection of U-Net architectures and loss functions for class imbalance},%
  pdfauthor={Valentyna Starodub and Mantas Lukosevicius},%
  pdfkeywords={age-related macular degeneration; color fundus retinography; biomedical imaging; lesion segmentation; U-Net; weighted binary cross-entropy},%
  pdfstartview=FitH,%
  bookmarks=true,%
  bookmarksopen=true,%
  breaklinks=true,%
  colorlinks=true,%
  linkcolor=refcolor,%
  anchorcolor=blue,%
  citecolor=refcolor,%
  filecolor=blue,%
  menucolor=blue,%
  urlcolor=refcolor,%
  pdfpagelabels]{hyperref} 
\else 
\usepackage[%
  breaklinks=true,%
  colorlinks=true,%
  linkcolor=blue,%
  anchorcolor=blue,%
  citecolor=blue,%
  filecolor=blue,%
  menucolor=blue,%
  urlcolor=blue]{hyperref} 
\fi 

\providecommand{\orcid}[1]{\href{https://orcid.org/#1}{\includegraphics[scale=0.5]{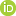}} }

\title{Surpassing state of the art on AMD area estimation from RGB fundus images through careful selection of U-Net architectures and loss functions for class imbalance}

\author{Valentyna Starodub$^{\dagger}$\orcid{0009-0007-0141-2233} \and Mantas Luko\v{s}evi\v{c}ius$^{\ddagger}$\orcid{0000-0001-7963-285X}\vspace{5pt}
}

\date{{\small Faculty of Informatics, Kaunas University of Technology,\\
LT-51368 Kaunas, Lithuania\\ 
$\dagger$ vlntnstarodub@gmail.com, \quad $\ddagger$ mantas.lukosevicius@ktu.lt}\\[2ex]%
\today}

\begin{document}

\maketitle

\begin{abstract}
Age-related macular degeneration (AMD) is one of the leading causes of irreversible vision impairment in people over the age of 60. This research focuses on semantic segmentation for AMD lesion detection in RGB fundus images, a non-invasive and cost-effective imaging technique. The results of the ADAM challenge -- the most comprehensive AMD detection from RGB fundus images research competition and open dataset to date -- serve as a benchmark for our evaluation.
Taking the U-Net connectivity as a base of our framework, we evaluate and compare several approaches to improve the segmentation model's architecture and training pipeline, including pre-processing techniques, encoder (backbone) deep network types of varying complexity, and specialized loss functions to mitigate class imbalances on image and pixel levels. The main outcome of this research is the final configuration of the AMD detection framework, which outperforms all the prior ADAM challenge submissions on the multi-class segmentation of different AMD lesion types in non-invasive RGB fundus images.
The source code used to conduct the experiments presented in this paper is made freely available.
\end{abstract}


\section{Introduction}
This research aims to investigate the application of machine learning methods to the field of ophthalmology, particularly automatic methods to detect age-related macular degeneration. 

Age-related macular degeneration (AMD) is a progressive eye disease that damages a central portion of the retina responsible for sharp central vision ~\cite{nihAgeRelatedMacular}. It is a leading cause of irreversible vision impairment in those over the age of 60 years in developed countries, affecting 200 million people worldwide. Early detection is crucial, as a timely assessment of the size and location of the lesion can guide effective treatment. However, diagnosis might be highly complicated, since in the early and intermediate stages, AMD is asymptomatic. In addition, easier-to-evaluate diagnostic methods, such as OCT or fluorescein angiography, are invasive, expensive, and time-consuming.

Therefore, in this research, we evaluate the approaches to improve the performance of the deep learning training/evaluation pipeline for AMD lesion detection in non-invasively registered RGB fundus images, including the choice of segmentation architectures and loss functions, to achieve improvement over previous benchmarks. We provide the source code for the research at \url{https://github.com/vlntn-starodub/AMD-lesion-segmentation}.

\section{Related work}
This section provides an overview of age-related macular degeneration and deep learning techniques for image segmentation, with a focus on retinal images and AMD detection using RGB fundus images.

\subsection{Age-related macular degeneration}

\begin{figure}[ht!]
\centering
\includegraphics[width=0.75\textwidth]{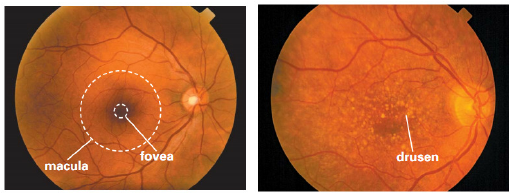}
\caption[A healthy retina (left) vs. AMD with drusen (right)]{\centering \small A healthy retina (left) and an AMD-affected retina with yellow drusen (right)~\cite{bhuiyan2014review}}
\label{fig:AMD_example}
\end{figure}

Age-related macular degeneration (AMD) is a progressive retinal disease that affects the macula (see Fig.~\ref{fig:AMD_example}), the central part of the retina responsible for sharp vision, which is critical for reading, driving, and facial recognition. Nearly 200 million people globally are affected by AMD~\cite{ruia2023macular}.

Early and intermediate AMD are typically asymptomatic and detectable only by eye examination. However, late AMD leads to vision loss and is the leading cause of irreversible visual impairment. The chance of late AMD-related visual loss can be reduced by taking certain actions when AMD is detected in earlier stages~\cite{NationalEyeInstitute}; therefore, early diagnosis is crucial. 

The diagnosis of AMD often involves imaging techniques such as optical coherence tomography (OCT), fundus autofluorescence (FAF), and contrast-enhanced RGB imaging. 
However, these techniques require expensive equipment, have limited availability, or are costly and invasive to the patient.

Non-contrast RGB fundus imaging (Color Fundus Retinography) is widely accessible and non-invasive, using red, green, and blue channels to visualize macular structures. It helps identify AMD markers such as drusen and pigment changes. However, interpretation can be challenging, motivating the development of automated detection methods using RGB images.
 
\subsection{Detection of eye diseases using RGB fundus images}
In the segmentation task using RGB fundus images, the techniques typically align with general approaches in semantic segmentation. Reviews from 2020 to 2023 highlight several key methods and architectures that are commonly used in biomedical image analysis, including encoder-decoder architectures such as U-Net and fully convolutional networks, skip connections, and dilated convolutions \cite{yu2023techniques, ulku2022survey, sohail2022systematic}. These models provide strong performance for high-precision tasks such as lesion segmentation and are promising for the application to the detection of AMD.

Based on the analysis of related works and their proven effectiveness in biomedical image segmentation \cite{goutam2022comprehensive, li2021applications, hoque2021deep}, U-Net \cite{ronneberger2015u} is chosen as the main architecture used in this research. It consists of an encoder-decoder structure with symmetric skip connections that enable both precise localization and context awareness, which is beneficial for generating detailed segmentation maps for AMD lesions in RGB fundus images.

Furthermore, EfficientNet \cite{tan2019efficientnet} is used as the backbone in this research, initialized with ImageNet pretrained weights \cite{deng2009imagenet}, which allows the model to benefit from transfer learning for improved generalization and faster convergence, important in cases with limited training data, as in this study. EfficientNet introduces compound scaling, which enhances efficiency and adaptability and reduces computational complexity through its use of depthwise separable convolutions.

Using EfficientNet as the backbone, U-Net shows improved feature representation and more efficient computation. For AMD segmentation in RGB fundus images, EfficientNet-B0, B1, and B2 can be the most suitable to balance accuracy and efficiency.

\subsection{Detection of age-related macular degeneration using RGB fundus images}\label{dataset}

Most research on AMD segmentation and detection focuses on OCT or contrast-enhanced RGB fundus images, which do not apply to this research. However, some studies align more closely with our task. The study \cite{ali2024amdnet23} introduces a deep learning–based approach using the hybrid AMDNet23 model, combining CNN and LSTM with preprocessing techniques such as gamma correction and CLAHE to enhance early detection and demonstrate the potential of hybrid models in medical imaging. Another study, \textit{Automated age-related macular degeneration area estimation—first results} \cite{pevciulis2021automated}, focuses on AMD detection from RGB fundus images using a dataset from the Lithuanian University of Health Sciences. It uses a custom classifier and four segmentation architectures, including U-Net, where segmentation is performed only if AMD is detected. The system demonstrates high accuracy in both classification and segmentation.

The most comprehensive AMD detection research using RGB fundus images is the ADAM Challenge at ISBI 2020 \cite{fang2022adam}, which aimed to improve algorithms for AMD diagnosis and lesion segmentation. Among the four tasks of the challenge, “Detection and Segmentation of Lesions” is most relevant to our research, involving the detection and pixel-wise segmentation of drusen, exudate, hemorrhage, scars, and other lesions. 

As a result of the challenge, several segmentation models for the detection of AMD from retinal images were developed and evaluated. In particular, the most popular architectures were U-Net, FPN, and DeepLab-v3. The U-Net models with EfﬁcientNet or Residual blocks as encoders showed the best performance in the lesion segmentation and detection part of the task. 

\begin{figure}[ht]
\centering
\includegraphics[width=\textwidth]{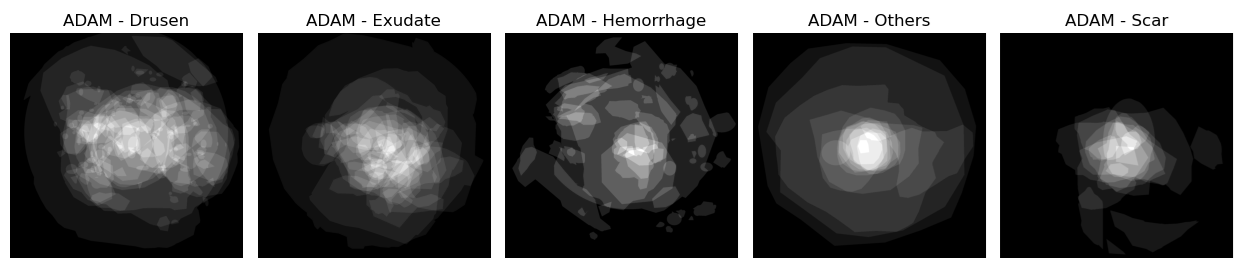}
\caption[Average of the masks over different types of lesions for ADAM dataset and over all the Lithuanian University of Health Sciences data.]{\centering \small Average of the masks over different types of lesions for ADAM dataset (the first five from the left)}
\label{fig:avg_masks}
\end{figure}

The dataset includes 1200 high-quality RGB fundus images split into equal training, validation, and test sets of 400 images each, as well as pixel-wise segmentation masks of five types of lesions (drusen, exudate, hemorrhage, scar, and others), shown in Fig.~\ref{fig:avg_masks}. 
118, 123, and 99 images, available in training, validation, and test datasets, respectively, have a segmentation mask for any out of the 5 lesions; some images have more than one corresponding lesion. The datasets exhibit a significant class imbalance, as images without lesions are presented in a much higher number (71\% of the training dataset) than those with segmentation masks. The sparseness leads to the model becoming biased toward the majority class, predicting the absence of the lesions, and reducing the performance on lesion detection.  

Furthermore, the ADAM challenge provides a description of the evaluation setup for detection models with the focus on tracking the Dice coefficient for segmentation and the F1 score for classification. This aligns with the commonly used metrics, described in the surveys on image segmentation, such as~\cite{ulku2022survey}.

The Dice coefficient measures the overlap between the predicted and reference masks. It is calculated on the pixel level and takes into account only the images that have regions of interest in them. The F1 score is used for the evaluation of the classification task, which assesses whether the region of interest is identified in an image, irrespective of precise pixel-level segmentation. The F1 score is calculated for all images. 

The evaluation of the model is based on a weighted combination of the Dice coefficient and the F1 score.

\begin{equation}
R = 0.4 \cdot \text{F1} + 0.6 \cdot \text{Dice}
\label{eq:rank}
\end{equation}

This weighting prioritizes segmentation performance, as it often has greater clinical significance with information about the size and shape of the lesion, which are critical for diagnosis and treatment planning.

The ADAM challenge dataset is used as the main input for this research. Therefore, the results obtained from the ADAM challenge are also used as a baseline for comparing the results of this research.

\section{Methods}
The research addresses semantic segmentation and detection of AMD lesions in RGB fundus images using a segmentation-based classification approach, with a focus on optimizing the training pipeline.

\subsection{Semantic segmentation and detection}

Since diagnosing AMD requires identifying not only the presence of lesions but also their size and location, this research explores the detection of lesions in RGB fundus images using binary classification and segmentation, both of which are essential for clinical decision-making. Providing both classification labels and segmentation masks makes the system more flexible to real-life demands: while classification indicates disease presence, segmentation provides spatial details for monitoring and treatment planning.

In this research, segmentation-based classification is used.
A segmentation model is trained on both AMD and non-AMD images. During inference, the model generates binary masks for all images. Classification is derived from the predicted mask: if any lesion pixel is detected, the image is classified as AMD. The final output includes a segmentation mask and a scalar label indicating the presence of AMD, based on the predicted mask.

This approach improves detection accuracy with detailed analysis from the segmentation model for the classification decision, potentially reducing false negatives. It also provides a simpler pipeline with a single model for both tasks. 

\subsection{Data preprocessing and loading pipeline}

As mentioned in the description of the dataset in Section~\ref{dataset}, in this research, some input images do not contain ROI, only background pixels, while others contain ROI and therefore have a corresponding ground-truth mask of the object of interest.

\begin{figure}[ht!] 
\centering
\includegraphics[width=1\textwidth]{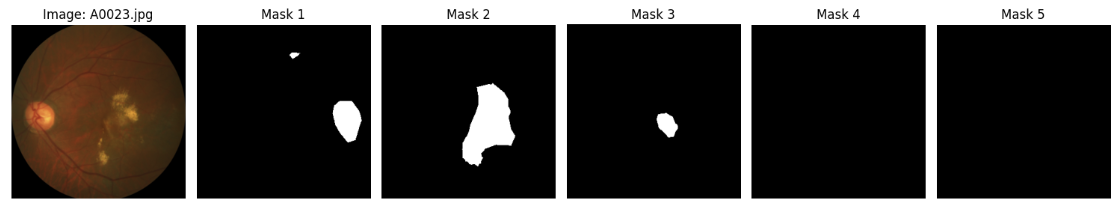}
\caption[Input image and multi-channel ground truth mask]{\centering \small Input image and multi-channel ground truth mask}
\label{fig:image_mask_ex}
\end{figure}

The multiclass binary segmentation task is considered: for each image, the pipeline constructs a multi-channel mask, where each channel represents a specific lesion type. Predictions are made for each lesion type separately. Missing masks for any lesion type are handled by filling the corresponding channel with zeros, indicating the absence of lesions of that type. The final multi-channel mask contains segmentation masks for all lesion types for a given image, illustrated in Fig.~\ref{fig:image_mask_ex}. 

\begin{figure}[h!]
\centering
\includegraphics[width=1\textwidth]{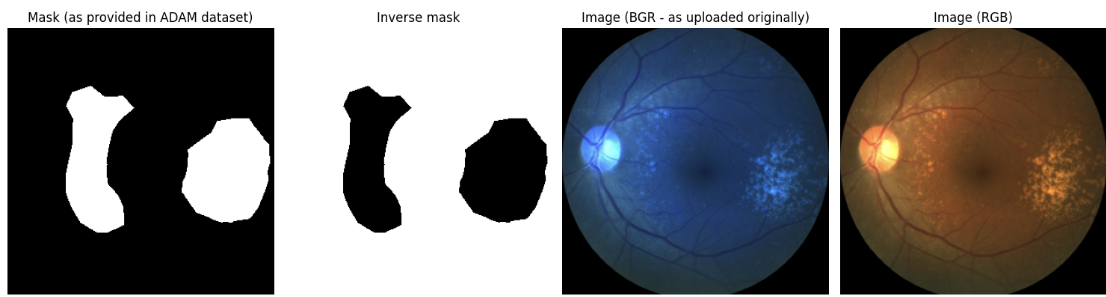}
\caption[Original and modified in data loading input image and ground truth mask]{\centering \small Original and modified in data loading input image and ground truth mask}
\label{fig:data_transform}
\end{figure}

Masks are loaded as grayscale images, binarized by converting pixel values of 255 to 1, and inverted to follow the convention of representing foreground as 1 and background as 0. Original and processed masks are shown in Fig.~\ref{fig:data_transform}.

All images and their corresponding masks are resized. The optimal resolution was determined by evaluating the performance of the model across multiple input sizes (\(640 \times 640\), \(320 \times 320\), and \(160 \times 160\)). Although lesion type and metric have an impact on the choice of optimal image size, it was decided to use $320\times 320$ size: this set-up demonstrated the highest Rank (Eq.~\ref{eq:rank}) compared to all other configurations for the majority of lesion types. Moreover, with limited computational resources, it still allowed for a large batch size in training, which had a major impact on performance improvement. Therefore, a resolution of \(320 \times 320\) was selected to balance the accuracy of the model with the computational efficiency during training and evaluation.

To enhance generalization, we applied on-the-fly data augmentation. Techniques included small rotations, cropping, scaling, and brightness/contrast adjustments to simulate variations in orientation, positioning, and lighting. Parameters were selected through tuning, and identical transformations were applied to both images and masks to ensure consistency.

The final sample of the dataset for each input image consists of the processed image tensor and the multi-channel mask tensor.

\subsection{Model architecture and initialization}

In this work, a modified U-Net model architecture from Segmentation Models Pytorch \cite{Iakubovskii} with an EfficientNet-based encoder (timm implementation of the PyTorch image models library \cite{rw2019timm}) is used. Both the impact of encoder depth and the necessity of using an encoder were evaluated. The encoder is pre-trained on the ImageNet dataset \cite{deng2009imagenet}. 

The model accepts three-channel (RGB) images and five-channel (one for each lesion) ground-truth masks and outputs a single-channel binary mask to indicate the lesion.

For each lesion type, a separate model is trained, optimized, validated, and tested, although the architecture and initial training and evaluation setup remain the same for all lesions.

\subsection{Loss functions for semantic segmentation}

As mentioned in Section~\ref{dataset}, the dataset shows a severe class imbalance between the foreground and background, due to both the limited number of ROI-containing images and the sparse foreground regions. Standard cross-entropy loss fails to adequately handle such an imbalance. To address this, we compare specialized loss functions -- such as weighted binary cross-entropy, Dice, Focal, and Tversky loss -- that focus on foreground prediction by applying penalty strategies. 

The weighted binary cross-entropy assigns class-specific weights to address this issue \cite{azad2023loss}. For sparse targets, where the foreground is underrepresented, the positive class weight is increased to improve foreground classification, making it beneficial for handling class imbalance in segmentation tasks.

The weights can be calculated as the proportion between the number of foreground and background pixels. The positive weight for each lesion type \( i \) is computed as:

\begin{equation}
\text{pos\_weights}[i] =
\begin{cases} 
\frac{\text{num\_neg}[i]}{\text{num\_pos}[i]}, & \text{if } \text{num\_pos}[i] > 0, \\
0, & \text{if } \text{num\_pos}[i] = 0, 
\end{cases}
\label{eq:pos_weights}
\end{equation}

\[
\text{num\_pos}[i] = \sum_{b=1}^{B} \sum_{x,y} \mathbbm{1}\{m_{b,i}(x,y) = 1\}
\]
\[
\text{num\_neg}[i] = \sum_{b=1}^{B} \sum_{x,y} \mathbbm{1}\{m_{b,i}(x,y) = 0\},\]
where \( T \) is the number of lesion types, \( B \) is the batch size, and \( m_{b,i}(x,y) \) represents the mask value for lesion type \( i \) at pixel location \( (x, y) \) in image \( b \).

Focal loss extends weighted cross-entropy loss with the addition of focusing parameters, which dynamically influence the impact of predicted probability \cite{azad2023loss}. It downweights easy examples and reduces the relative loss for well-classified examples, allowing the model to concentrate on difficult cases. 

The focusing parameter must be tuned. The values of the weighted parameters can be computed by normalizing the positive weights of the weighted BCE \( \text{pos\_weights}[i] \) (Equation~\ref{eq:pos_weights}). For each lesion type \( i \), the alpha is given by:

\begin{equation}
 \alpha[i] = \frac{\text{pos\_weights}[i]}{\sum_{j=1}^{T} \text{pos\_weights}[j]},
 \label{eq:alpha}
\end{equation}
where \( \text{pos\_weights}[i] \) is the positive weight for lesion type \( i \);  \( T \) is the total number of lesion types.

Focal loss is effective when a significant proportion of the images are part of the background. However, its dynamic reweighting might struggle in cases of severe imbalance with sparse, irregular positive classes.

Dice loss, derived from the Dice coefficient, optimizes the overlap between the predicted and target regions and is widely used in segmentation \cite{azad2023loss}. However, it can be unstable with sparse foregrounds (as seen in Fig.~\ref{fig:output}) due to its sensitivity to small prediction errors. 

Tversky loss, a generalization of Dice, introduces adjustable parameters to better handle class imbalance by weighting false positives and false negatives \cite{azad2023loss}. Although more flexible, it still relies on overlap and may struggle with irregular foregrounds. Another disadvantage is the need to tune the parameters rather than having them determined through calculations.

\subsection{Training and evaluation pipeline}
A model for each lesion type is trained independently over 100 epochs. To reduce overfitting, the best-performing model (based on validation Rank, Eq.~\ref{eq:rank}) is saved whenever performance improves. This allows recovery and testing of the best model without retraining. 

For inference, inputs are preprocessed consistently with training, and the model outputs are post-processed (thresholding) to generate segmentation masks and classification labels. The validation metrics -- the Dice coefficient for segmentation, the F1 score for the detection of the presence of lesions, and the Rank (\ref{eq:rank}) -- are computed separately for each lesion type.

\begin{figure}[h]
\centering
\includegraphics[width=\textwidth]{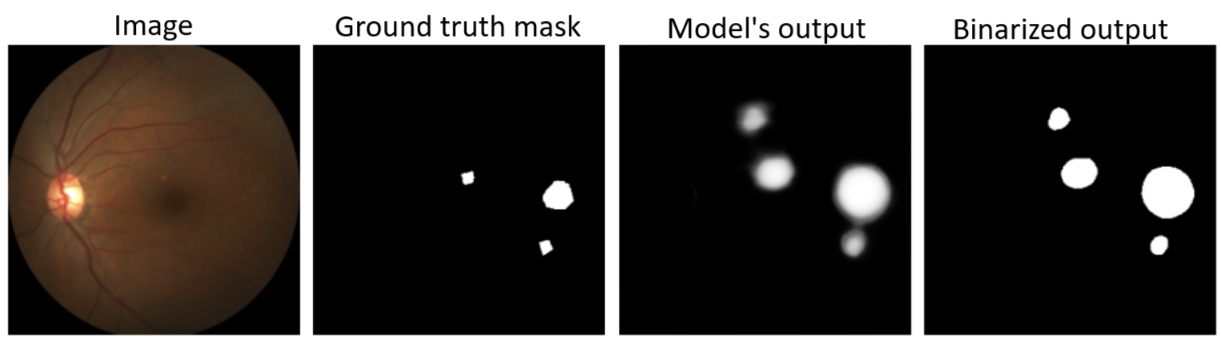}
\caption[Input image, ground truth mask, output of segmentation model and its binarization (threshold = 0.5) ]{\centering \small Input image, ground truth mask, output of segmentation model and its binarization (threshold = 0.5) }
\label{fig:output}
\end{figure}

Evaluation combines quantitative metrics (F1, Dice scores) and qualitative visual inspection, ensuring that masks accurately highlight target regions (as shown in Fig.~\ref{fig:output}).

Framework configurations were selected based on validation set performance. The test set was reserved for evaluating the final model and comparing it with those of the ADAM challenge.

\section{Results}
Several approaches to improve segmentation performance were evaluated, including variations in encoder complexity, attention mechanisms, loss functions, and dropout rates. However, only the choice of encoder and loss function significantly impacted training outcomes and are therefore discussed in detail. The main outcome is the final configuration of the AMD detection framework and its comparison with the ADAM challenge results.

\subsection{Evaluation of the encoder choice impact}

The U-Net was chosen due to its common use in medical segmentation, highlighted in the literature review. Initially, a traditional U-Net implementation was used. However, due to the simplicity of the manually created architecture, this approach was insufficient: for all lesion types, the Dice score on the validation dataset was near 0. 

Therefore, we switch to a more optimized U-Net architecture from the Segmentation Models Pytorch (SMP) library~\cite{Iakubovskii}, which supports pre-trained encoders for improved feature extraction and faster convergence. EfficientNetB0 and EfficientNetB2 encoders were evaluated, suited for low- and medium-complexity tasks, respectively. Additionally, the timm library version of the encoders was tested, offering better pretrained weights due to enhanced optimizers, augmentations, and regularization during pretraining.

For a comprehensive comparison across all lesion types, average and weighted average metrics were used, the latter weighted by the number of images per lesion type in the validation set.

The results of the encoder comparison are provided in Table~\ref{tab:encoders}.

\begin{table}[h!]
\caption{Metrics for different EfficientNet encoders}
\label{tab:encoders}
\centering
\begin{tabular}{l l r r r}
\toprule
\textbf{Lesion} &\textbf{Metric} &\textbf{B0} &\textbf{B0 (timm)} &\textbf{B2 (timm)} \\ 
\midrule
\multirow{3}{*}{Drusen} &Dice &0.4872 &0.4853 &\textbf{0.5017}  \\
&F1 &0.8050 &0.8050 &\textbf{0.8275} \\ 
&Rank &0.6143 &0.6132 &\textbf{0.6320} \\
\midrule
\multirow{3}{*}{Exudate} &Dice &0.5970 &\textbf{0.6082} &0.5628  \\ 
&F1 &0.6850 &0.7675 &\textbf{0.7875} \\ 
&Rank &0.6322 &\textbf{0.6719} &0.6527 \\
\midrule
\multirow{3}{*}{Hemorrhage} &Dice &0.2554 &\textbf{0.3457} &0.3419 \\  
&F1 &\textbf{0.9575} &\textbf{0.9575} &0.9400 \\ 
&Rank &0.5362 &\textbf{0.5904} &0.5811 \\ 
\midrule
\multirow{3}{*}{Other} &Dice &0.2992 &0.3018 &\textbf{0.3355}  \\ 
&F1 &0.8925 &0.6800 &\textbf{0.9825} \\ 
&Rank &0.5365 &0.4531 &\textbf{0.5943} \\ 
\midrule
\multirow{3}{*}{Scar} &Dice &0.4821 &\textbf{0.6650} &0.5313  \\ 
&F1 &0.9475 &0.8125 &\textbf{0.9725} \\ 
&Rank &0.6683 &\textbf{0.7240} &0.7077 \\ 
\midrule
\multirow{3}{*}{\textbf{Average}} &Dice &0.4242 &\textbf{0.4812} &0.4546  \\ 
&F1 &0.8575 &0.8045 &\textbf{0.9020} \\ 
&\textbf{Rank} &0.5975 &0.6105 &\textbf{0.6336} \\ 
\midrule
\multirow{3}{*}{\shortstack[l]{\textbf{Weighted}\\ \textbf{average}}} &Dice &0.4484 &\textbf{0.4705} &0.4651  \\ 
&F1 &0.8124 &0.8006 &\textbf{0.8629} \\ 
&\textbf{Rank} &0.5940 &0.6025 &\textbf{0.6242} \\ 
\bottomrule
\end{tabular}
\end{table}

The evaluation results show that using the timm library significantly improves the performance of the model: the timm-based EfficientNetB0 consistently outperforms the standard version in the Dice and Rank metrics for exudate, hemorrhage, and scar lesions, and achieves a higher F1 score for exudates. It also shows better average and weighted average Dice and Rank scores for all lesion types. This highlights the importance of advanced pretraining strategies of encoders for accurate lesion segmentation.

When comparing the usage of architectural complexities, EfficientNetB2 outperforms EfficientNetB0 in classification, achieving higher F1 scores for all lesions except hemorrhage. However, in segmentation, it performs better only for drusen and “other” lesions, indicating that although EfficientNetB2 is effective at detecting lesions, it struggles with precise boundary segmentation.

Despite these segmentation limitations, EfficientNetB2 maintains strong overall performance: both average Rank scores are the highest among all encoders, and it achieves the second-best Dice score, showing strong combined performance in both classification and segmentation tasks.

\subsection{Evaluation of the loss function impact}\label{loss}
Four loss functions commonly used in semantic segmentation tasks with class imbalance were analyzed: weighted binary cross-entropy loss (the weights for the positive class (foreground pixels) calculated based on the proportion of foreground and background pixels), Dice loss, Tversky loss, and Focal loss.

For Tversky and Focal loss, tuning is the primary way to find the optimal parameters.

 Tversky loss was tested with a parameter $\alpha$ of 0.1, 0.2, 0.3, 0.4, $\beta = 1-\alpha$ -- corresponding to penalizing false negatives more to help detect foreground effectively.

 \begin{figure}[h]
\centering
\includegraphics[width=1\textwidth]{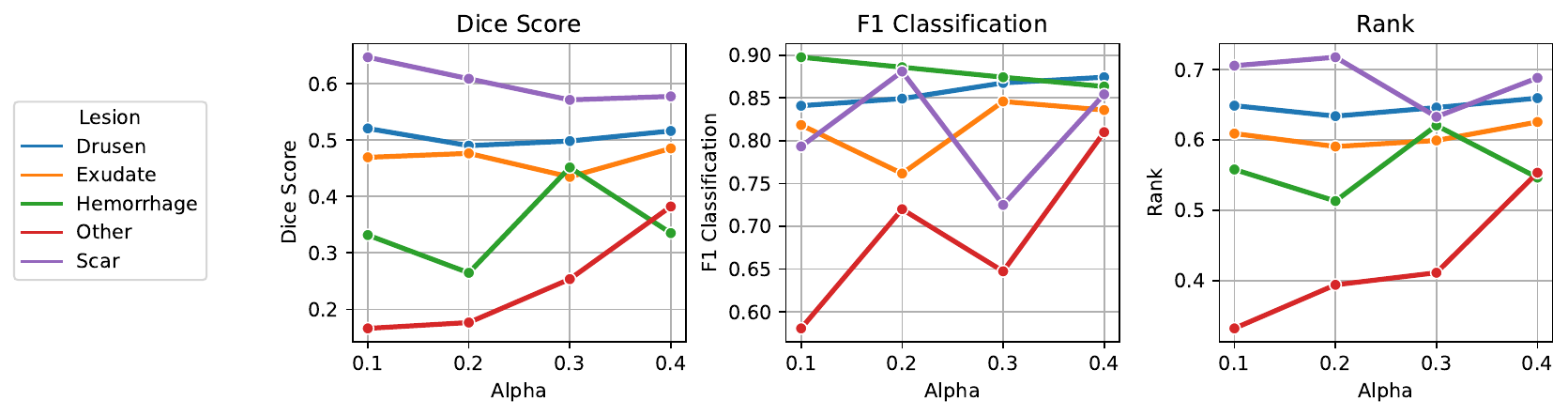}
\caption[Results of parameter tuning for Tversky loss; $\alpha$ equal to 0.1, 0.2, 0.3, 0.4, $\beta = 1-\alpha$]{\centering \small Results of parameter tuning for Tversky loss; $\alpha$ equal to 0.1, 0.2, 0.3, 0.4, $\beta = 1-\alpha$}
\label{fig:tversky}
\end{figure}

Focal loss was tested with the parameter $\alpha$ calculated based on the proportion of fore- and background pixels (\ref{eq:alpha}), and $\gamma$ equal to 1 (no focus on underrepresented foreground), 2 (moderate focus on hard-to-classify pixels) or 3 (strong focus on misclassified regions, useful for small foreground objects).

\begin{figure}[h]
\centering
\includegraphics[width=1\textwidth]{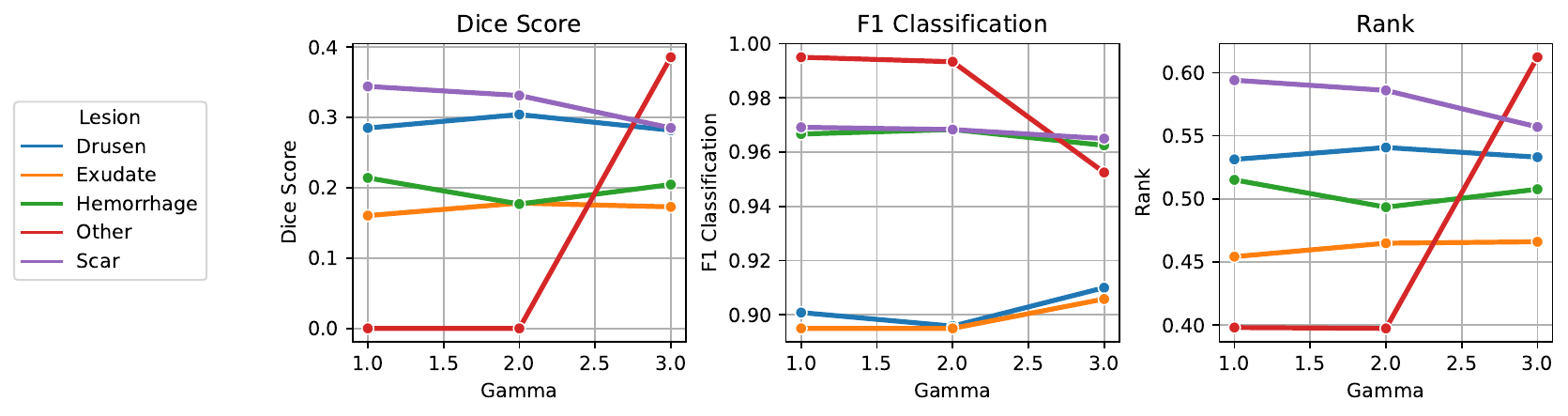}
\caption[Results of parameter tuning for Focal loss; $\alpha$ calculated based on the proportion of fore- and background pixels (\ref{eq:alpha}), $\gamma$ is equal to 1, 2 or 3]{\centering \small Results of parameter tuning for Focal loss; $\alpha$ calculated based on the proportion of fore- and background pixels (\ref{eq:alpha}), $\gamma$ is equal to 1, 2 or 3}
\label{fig:Focal}
\end{figure}

As shown in Figs.~\ref{fig:tversky} and \ref{fig:Focal}, performance is different between the lesion types depending on the parameter values, indicating that some lesions may need more aggressive weight balancing to address the class imbalance. The optimal parameters for each lesion type are listed in Table~\ref{tab:loss_parameters}.

 \begin{table}[ht]
\caption{Optimal parameters for Tversky and Focal loss}
\centering
\begin{tabular}{l r r r r r} 
\toprule
\textbf{Loss} & \textbf{Drusen} & \textbf{Exudate} & \textbf{Haemorrhage} & \textbf{Other} & \textbf{Scar}\\ 
\midrule
Tversky (\(\alpha\))& 0.4 & 0.4 & 0.3 & 0.4 &0.2 \\ 
Focal  (\(\gamma\)) & 2 & 3 & 1 & 3 & 1 \\
\bottomrule
\end{tabular}
\label{tab:loss_parameters}
\end{table}

\begin{figure}[h]
\centering
\includegraphics[width=\textwidth]{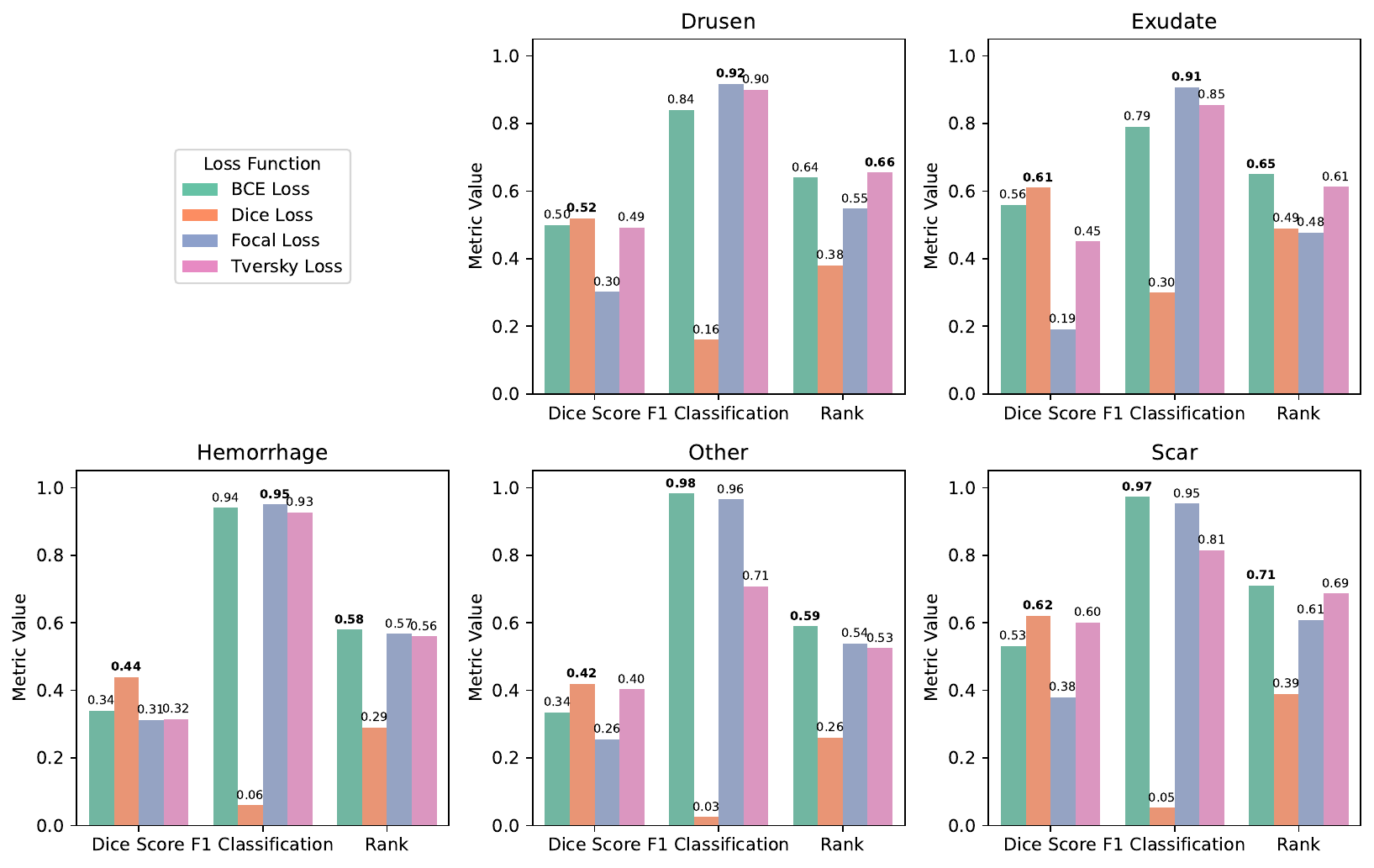}
\caption[Comparison of the results obtained after training with the usage of different loss functions]{\centering \small Comparison of the results obtained after training with the usage of different loss functions}
\label{fig:losses}
\end{figure}

Fig.~\ref{fig:losses} illustrates the results obtained using different loss functions.

Weighted binary cross-entropy and Tversky loss functions demonstrate the most consistent performance across both Dice and F1 scores. Although not always achieving the best performance in these metrics, their primary strength lies in balanced optimization of both segmentation and detection, shown in the highest Rank scores across all lesion types. This trade-off suggests that they avoid extensive focus on either image- or pixel-level accuracy.

The consistency of weighted binary cross-entropy loss stems from its simple handling of class imbalance by directly applying class-level weighting instead of compensating for skewed distributions by emphasizing harder-to-classify cases (as in Focal loss) or maximizing mask overlap (as in Dice loss).

The high performance of Tversky loss, a generalization of Dice loss, is better understood by examining the training with Dice loss.

\begin{figure}[h]
\centering
\includegraphics[width=\textwidth]{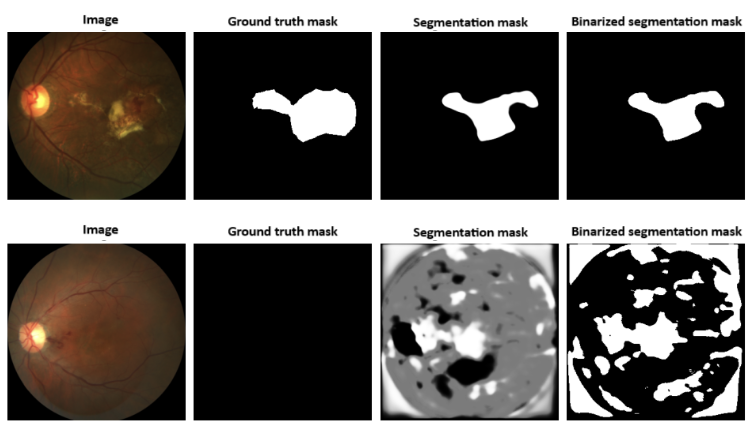}
\caption[Images, ground-truth and segmentation masks obtained from the model trained using Dice loss]{\centering \small Images, ground-truth and segmentation masks obtained from the model trained using Dice loss}
\label{fig:Dice_loss_ex}
\end{figure}

Dice loss appears to provide mixed results: it performs well in segmentation but has the lowest F1 scores for classification. However, the segmentation scores with Dice loss are misleading, as they are calculated only on samples with lesions. As illustrated in Fig.~\ref{fig:Dice_loss_ex}, Dice loss leads to oversegmentation: for images without the lesions, the model still returns a non-zero segmentation mask, which increases false positives and negatively impacts the classification. Dice loss fails to include images without lesions, making it less suitable for joint segmentation-classification tasks, but effective when all images contain lesions.

This leads to the use of Tversky loss: by introducing a penalty for false positives, while still keeping a higher penalty for false negatives to overcome foreground under-representation, we can overcome the shortcomings of Dice loss.

Focal loss showed the highest F1 score, while underperforming in pixel-wise segmentation, reflected in the lowest Dice coefficients. This can be explained by the more significant class imbalance on the pixel-wise level compared to the image-wise one. Therefore, Focal loss overcomes image-level imbalance by focusing on harder-to-classify instances. However, this strategy is not sufficient in the more complex case of an imbalance in segmentation.

From the comparison of the results, weighted BCE and Tversky loss showed the most consistent performance across all metrics and lesions, with the highest Rank values. The weighted BCE was selected for the final setup.  It demonstrated well-balanced performance in both segmentation and classification, while not requiring parameter tuning.

\subsection{Final configuration}
This section presents the final configuration based on commonly used techniques from the literature review and the analysis of the results from previous sections. This configuration addresses the challenges identified earlier and achieves the highest results for the automated AMD area estimation task.

Main parameters of the configuration:
\begin{itemize}
    \item input image and masks dimensions: $320\times 320$; 5-channel binarized masks, each channel corresponding to one lesion type (mask consists of pixels with value 0 if a lesion is not available);
    \item batch size: 32;
    \item model: U-Net with an EfficientNet-based encoder (timm PyTorch image models library implementation), pre-trained on the ImageNet dataset; dropout rate 0; includes batch normalization;
    \item optimizer: Adam optimizer; 
    \item learning rate: 0.001; learning rate scheduler: ReduceLROnPlateau (factor of 0.1 when validation loss stagnates for 25 epochs);
    \item loss function: weighted Binary Cross-Entropy loss function; weights (\ref{eq:pos_weights}) are assigned to a positive class for each lesion type (Table~\ref{table:weights});
    \item number of epochs: 100 (best model is saved during the training if the validation Rank has improved).
\end{itemize}
    
\begin{table}[ht]

\centering
\caption{Weights assigned to positive class in weighted Binary Cross-Entropy loss function}
\begin{tabular}{ l r r r r r}
\toprule
\textbf{Lesion type} & Drusen & Exudate & Haemorrhage & Other & Scar \\ 
\midrule
\textbf{Weight}  & 135  & 175  & 386  & 170  & 550  \\ 
\bottomrule
\end{tabular}
\label{table:weights}

\end{table}

Table~\ref{table:final} shows the results and their comparison to the ADAM challenge. “Best” and “2nd best” list the two highest reported values per metric and lesion, while “Best team” shows the top team's results based on the final ranking in the lesion segmentation task. The Rank metric, not provided in \cite{fang2022adam}, is calculated separately using~\ref{eq:rank}.

The best values for the Dice coefficient, the F1 score, and the Rank for each lesion are marked in bold.

\begin{table}[h!]
\centering
\caption{Performance metrics of our final configuration, and their comparison to results of the ADAM challenge}
\begin{tabular}{l l r r r r}
\toprule
\multirow{2.5}{*}{\textbf{Lesion}} & \multirow{2.5}{*}{\textbf{Metric}} & \multirow{2.5}{*}{\textbf{Our result}} & \multicolumn{3}{c}{\textbf{ADAM challenge results}} \\
\cmidrule(l){4-6}
& & &\textbf{Best} & \textbf{2nd best} & \textbf{Best team}\\ 
\midrule
\multirow{3}{*}{Drusen} &Dice &0.5102 &\textbf{0.5549} &0.4838 &0.4838 \\
&F1 &\textbf{0.7800} &0.6316 &0.5674 &0.6316 \\
&Rank &\textbf{0.6181} &0.5856 &0.5172 &0.5429 \\
\midrule
\multirow{3}{*}{Exudate} &Dice &\textbf{0.5846} &0.4337 &0.4154 &0.4154 \\
&F1 &\textbf{0.6900} &0.5688 &0.5581 &0.5688 \\
&Rank &\textbf{0.6268} &0.4877 &0.4725 &0.4768 \\
\midrule
\multirow{3}{*}{Haemorrhage} &Dice &0.3860 &\textbf{0.4303} &0.2400 &\textbf{0.4303} \\
&F1 &\textbf{0.9100} &0.8293 &0.7307 &0.7307 \\
&Rank &\textbf{0.5956} &0.5899 &0.4363 &0.5505 \\
\midrule
\multirow{3}{*}{Other} &Dice &0.3349 &\textbf{0.6906} &0.2852 &0.2852 \\
&F1 &\textbf{0.7450} &0.4724 &0.1818 &0.0714 \\
&Rank &0.4989 &\textbf{0.6033} &0.2438 &0.1997 \\
\midrule
\multirow{3}{*}{Scar} &Dice &\textbf{0.6747} &0.5807 &0.5639 &0.4051 \\
&F1 &\textbf{0.8950} &0.8511 &0.7273 &0.7027 \\
&Rank &\textbf{0.7628} &0.6889 &0.6293 &0.5241 \\
\midrule
\multirow{3}{*}{\textbf{Average}} &Dice &0.4981 &\textbf{0.5380} &0.3977 &0.4040 \\
&F1 &\textbf{0.8040} &0.6706 &0.5531 &0.5410 \\
&Rank &\textbf{0.6204} &0.5911 &0.4598 &0.4588 \\
\midrule
\multirow{3}{*}{\shortstack[l]{\textbf{Weighted}\\ \textbf{average}}} &Dice &0.4846 &\textbf{0.5185} &0.3886 &0.4200 \\
&F1 &\textbf{0.7721} &0.6253 &0.5334 &0.5402 \\
&Rank &\textbf{0.5996} &0.5613 &0.4465 &0.4681 \\
\bottomrule
\end{tabular}
\label{table:final}
\end{table}

\textbf{Drusen and hemorrhage:} Although the Dice score of our approach is slightly lower than the highest values achieved in the ADAM challenge (still outperforms the second-best Dice coefficient), the proposed method has a significantly higher F1 score. This suggests that while the prediction of the outline or location of the lesions might still be challenging, the model shows improvement in detecting the presence of the lesions. This also positively impacted the overall performance -- the Rank score is the highest among all the compared results. In addition, for drusen, the proposed model exceeds the result of the highest-ranked team for this lesion for all metrics; for hemorrhage, Dice is the only value lower than the one obtained by the highest-ranked team of the ADAM challenge.

\textbf{Exudate and scar:} The proposed method consistently outperforms all ADAM baselines across all metrics. Its Dice and F1 scores and, as a result, Rank are significantly higher, indicating that the model provides improved results in both segmenting and detecting the presence of this lesion type.

\textbf{“Other”:} In the category “other”, the proposed method provides exceptional results in terms of the F1 score, showing its ability to detect lesions. However, its Dice score lags behind the best ADAM model, suggesting the issues with the precision of predicting the outlines of the lesions; this negatively impacts the Rank metric. Despite this, the method performs better than the second-best model. Compared to the highest-ranking team, it shows a significant improvement, particularly in the F1 score, which rose from 0.07 to 0.98, showing its ability to maintain high performance on sparse data, where the top ADAM team struggled.

\textbf{Average and weighted average:} The results are similar to those observed with individual lesions: the proposed method surpasses all ADAM setups in the detection while providing only the second-best results in segmentation. However, when considering the combined performance across both tasks -- the Rank metric -- the framework showed improvement compared to all the results of the ADAM challenge.

Overall, the suggested setup is beneficial for detecting the presence of lesions, but the pixel-wise segmentation results, although competitive, can still be improved.

Another notable observation is that the proposed framework does not perform well only on a subset of lesion types while compromising performance on others, as was the case with some models of the ADAM challenge. For instance, the highest-ranked team model achieved strong results for certain lesion types (drusen or hemorrhage), but performed poorly on the “other” lesions. In contrast, the proposed framework maintains balanced, high performance across all lesion types.

In summary, the proposed method consistently outperforms the best or second-best results in the ADAM challenge for all types of lesions. The model achieves particularly strong results for exudate and scar lesion types, while maintaining competitive results for the other lesion types. Compared to the highest-ranked team in the lesion segmentation task, the proposed setup shows improved performance for all lesion types and metrics, except for the Dice score on hemorrhage lesions. These results suggest that the proposed method offers a promising alternative for lesion detection tasks.

\section{Discussion}

This study explored multiple strategies to enhance segmentation performance for AMD lesion detection in non-invasive RGB fundus images. Performance improvement was achieved through appropriate selection of the encoder -- particularly considering its pre-training strategy and complexity -- as well as the use of loss functions capable of mitigating class imbalance at both the pixel and image levels, since the key challenges of the segmentation task were the imbalance between foreground and background pixels and the difficulty in segmenting rare lesion types. 

The use of pre-trained encoders clearly improved segmentation and classification metrics across most lesion types. Moreover, the usage of the advanced pre-training strategies significantly enhanced model performance. It was also shown that deeper architectures (such as EfficientNetB2 compared to EfficientNetB0), while better in classification, showed mixed segmentation results, suggesting deeper encoders may be more effective in detecting lesion presence than capturing precise boundaries. Still, EfficientNetB2 achieved the highest overall performance and was included in the final configuration.

As for the choice of the loss function, out of four evaluated ones, weighted Binary Cross-Entropy (BCE) and Tversky showed the most consistent results across the metrics for all lesions, resulting in the best overall performance. Other functions -- Dice and Focal loss -- were able to demonstrate high results only in the segmentation or classification part of the task, respectively, neglecting, however, the other part. Focal loss achieved strong classification but weak segmentation, indicating its limitations in handling pixel-level imbalance. Dice loss, though effective in segmentation, led to a high amount of false positive predictions for the images without lesions, impairing classification performance. This prompted the usage of Tversky loss, which successfully addressed this issue by penalizing false positives more. Weighted BCE was ultimately chosen for its consistent performance and simplicity, requiring no parameter tuning.

The final configuration of the AMD lesion detection framework included a U-Net model with an ImageNet-pretrained EfficientNetB2 encoder, a weighted binary cross-entropy loss function with weights tuned to reflect the class imbalance of the dataset. The proposed method consistently outperforms previous submissions to the ADAM challenge on the multi-class segmentation of various AMD lesion types in non-invasive RGB fundus images, achieving state-of-the-art performance. Future work can focus on further improving lesion localisation and the precision of the lesion boundaries.

  \section{Conclusions}
This research investigated the application of semantic segmentation models for the automated age-related macular degeneration area estimation in non-invasively registered RGB fundus images, focusing on improving the accuracy of the semantic segmentation task and addressing associated challenges, such as class imbalance and sparse data.

The final configuration of the proposed AMD lesion detection framework, consisting of a U-Net model with an EfficientNet encoder pre-trained on ImageNet and a weighted binary cross-entropy loss function, achieved state-of-the-art performance in the semantic segmentation of AMD lesions. Comparative analysis against the best-performing methods from the ADAM challenge confirmed improved segmentation and detection accuracy.

The improvements over prior benchmarks demonstrate the applicability of the developed framework to automated, non-invasive diagnosis of AMD, and the insights provided in this research pertain to the broader medical image analysis field. 

The source code for the research is provided at \url{https://github.com/vlntn-starodub/AMD-lesion-segmentation}.

\section*{Funding sources}
This research did not receive any specific grant from funding agencies in the public, commercial, or not-for-profit sectors.

\section*{Author contributions}
\textbf{Valentyna Starodub:} Conceptualization, Methodology, Software, Validation, Investigation, Writing -- Original Draft, Visualization.
\textbf{Mantas Lukoševičius:} Conceptualization, Methodology, Resources, Writing -- Review \& Editing, Supervision.

 \bibliographystyle{IEEEtran} 
 \bibliography{bibliography}

\end{document}